%% file: main.tex
\documentclass[letterpaper]{article} % DO NOT CHANGE THIS
\usepackage{aaai2026}  % DO NOT CHANGE THIS
\usepackage{times}  % DO NOT CHANGE THIS
\usepackage{helvet}  % DO NOT CHANGE THIS
\usepackage{courier}  % DO NOT CHANGE THIS
\usepackage[hyphens]{url}  % DO NOT CHANGE THIS
\usepackage{graphicx} % DO NOT CHANGE THIS
\usepackage{natbib}  % DO NOT CHANGE THIS AND DO NOT ADD ANY OPTIONS TO IT
\usepackage{caption} % DO NOT CHANGE THIS AND DO NOT ADD ANY OPTIONS TO IT
\usepackage{algorithm}
\usepackage{algorithmic}
\usepackage{amsmath,amsfonts,bm}
\usepackage{multirow}
\usepackage{booktabs}

\usepackage{colortbl}
\usepackage{pifont}
\usepackage{bbding}   
\usepackage{fontawesome}  
\usepackage{color}
\usepackage[algo2e,ruled,noend,linesnumbered]{algorithm2e}
\usepackage{soul}
\usepackage{multirow}
\usepackage{makecell}

\usepackage{caption}
\usepackage{xcolor}
\definecolor{lavender}{RGB}{230, 230, 250}
\definecolor{LemonChiffon}{RGB}{240, 232, 240}
\definecolor{MistyRose}{RGB}{255, 228, 225}
\definecolor{yellow}{RGB}{241, 205, 122}
\usepackage{rotating}  

\usepackage{pifont}

\usepackage{bm}
\usepackage[misc]{ifsym}

\usepackage{textcomp}  
\usepackage{scalerel}

\usepackage{tikz}

\newcommand{\decrease}[1]{
	{\fontsize{7pt}{0.5em}\selectfont\color{red!100}{$\downarrow$~{#1}}}
}

\usepackage{newfloat}
\usepackage{listings}
\DeclareCaptionStyle{ruled}{labelfont=normalfont,labelsep=colon,strut=off} % DO NOT CHANGE THIS
\floatstyle{ruled}
\newfloat{listing}{tb}{lst}{}
\floatname{listing}{Listing}
\title{Q Cache: Visual Attention is Valuable in Less than Half of Decode Layers \\ for Multimodal Large Language Model}
\author {
    Jiedong Zhuang\textsuperscript{\rm 1,\rm 2},
    Lu Lu\textsuperscript{\rm 2},
    Ming Dai\textsuperscript{\rm 1},
    Rui Hu\textsuperscript{\rm 1},
    Jian Chen\textsuperscript{\rm 2},
    Qiang Liu\textsuperscript{\rm 2},
    Haoji Hu\textsuperscript{\rm 1}\corresponding
}
\affiliations {
    \textsuperscript{\rm 1}Zhejiang University\\
    \textsuperscript{\rm 2}Alibaba Cloud Computing\\
    zhuangjiedong@zju.edu.cn, ll200214@alibaba-inc.com, mingdai997@outlook.com, rickhu@zju.edu.cn, j.chen@alibaba-inc.com,
    yf.yifeng@alibaba-inc.com, haoji\_hu@zju.edu.cn
}

\usepackage{bibentry}
\begin{document}

\maketitle
\input{sec/abstract}
\input{sec/introduction}

\input{sec/related}
\input{sec/method}

\input{sec/experiments}
\input{sec/conclusion}

%\bibliography{aaai2026}
% \bibliography{aaai2026}

\end{document}

%% file: sec/abstract.tex
\begin{abstract}
Multimodal large language models (MLLMs) are plagued by exorbitant inference costs attributable to the profusion of visual tokens within the vision encoder. The redundant visual tokens engenders a substantial computational load and key-value (KV) cache footprint bottleneck. Existing approaches focus on \textbf{token-wise} optimization, leveraging diverse intricate token pruning techniques to eliminate non-crucial visual tokens. Nevertheless, these methods often unavoidably undermine the integrity of the KV cache, resulting in failures in long-text generation tasks.
To this end, we conduct an in-depth investigation towards the attention mechanism of the model from a new perspective, and discern that \textbf{attention within more than half of all decode layers are semantic similar}. Upon this finding, we contend that the attention in certain layers can be streamlined by inheriting the attention from their preceding layers. Consequently, we propose \textbf{Lazy Attention}, an efficient attention mechanism that enables cross-layer sharing of similar attention patterns.
It ingeniously reduces \textbf{layer-wise} redundant computation in attention. In Lazy Attention, we develop a novel layer-shared cache, \textbf{Q Cache}, tailored for MLLMs, which facilitates the reuse of queries across adjacent layers.
In particular, Q Cache is lightweight and fully compatible with existing inference frameworks, including Flash Attention and KV cache. 
Additionally, our method is highly flexible as it is orthogonal to existing token-wise techniques and can be deployed independently or combined with token pruning approaches. Empirical evaluations on multiple benchmarks demonstrate that our method can reduce KV cache usage by over \textbf{35\%} and achieve \textbf{1.5$\times$} throughput improvement, while sacrificing only approximately 1\% of performance on various MLLMs. Compared with SOTA token-wise methods, our technique achieves superior accuracy preservation.
\end{abstract}

%% file: sec/introduction.tex
\section{
Introduction}
\label{sec:intro}

\begin{figure}[t]
\centering
\includegraphics[width=0.47\textwidth]{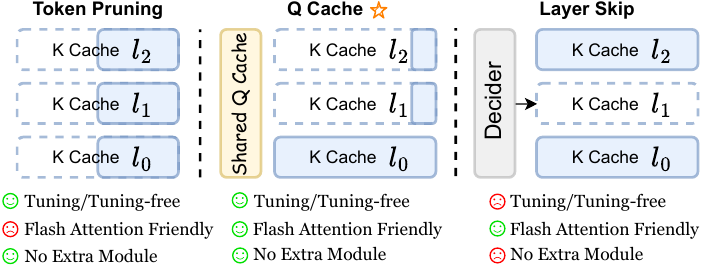}
\caption{A brief comparison on K Cache in varied decode layers of a MLLM. We introduce a new technique Q Cache, which significantly reduces the K Cache memory footprint. Compared to other methods, it is more flexible, plug-and-play, and compatible with Flash Attention.
}
\end{figure}

Multimodal Large Language Models (MLLMs) \cite{llava1.5,llava-next} have exhibited remarkable capabilities in various multimodal tasks, such as visual question answering \cite{SQA}, visual reasoning \cite{mme} and visual grounding \cite{c3vg,falip}. Nonetheless, due to the inherent nature of their paradigms, some modalities, e.g., images and videos \cite{video-llava,video-llama}, generate a substantial number of visual tokens through vision encoders \cite{CLIP}. These numerous tokens significantly affect the inference speed of Transformer-based \cite{attention} MLLMs, as the computational complexity of attention scales quadratically with the length of token sequence. 
Moreover, visual contexts usually introduce excessive key-value (KV) cache \cite{kvcache}, significantly limiting generative efficiency due to memory bottlenecks.

Against this backdrop, a plethora of acceleration optimization methods based on token pruning have emerged in the realm of MLLMs \cite{fastv,VTW,st3,pyramiddrop,sparsevlm}, which primarily focus on reducing visual redundancy in a \textit{token-wise} manner. Most of them rely on the magnitude of the values corresponding to each visual token in the attention maps to determine whether to retain or discard that token. However, this attention-based strategy requires real-time access to the attention matrix, rendering it incompatible with Flash Attention \cite{flashattention2}. Moreover, these methods \cite{fastv,VTW,st3,pyramiddrop,sparsevlm} propose to discard visual tokens during the prefill phase, which simultaneously compromises the integrity of the KV cache \cite{kvcache}, making the optimized models struggle to maintain stable performance in multi-turn dialogue scenarios \cite{can}. In this regard, we question whether the attention in all these layers are truly necessary? Can we approach optimization from a layer-wise perspective to streamline the bulky attention at a cross-layer granularity?

\begin{figure}[t]
\centering
\includegraphics[width=0.45\textwidth]{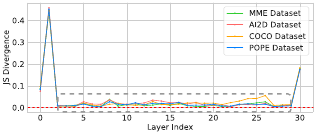}
\caption{Jensen-Shannon (JS) divergence of the attention scores between each layer and its adjacent layer below it. 
The gray dashed boxes highlight the regions with values close to 0, where the attention score distributions in the two layers are nearly identical, suggesting that the adjacent decode layers share highly similar attention patterns.
}
\label{attn_js}
\end{figure}

% To gain deeper insights, 
To address above questions, we begin by conducting an in-depth investigation of existing attention mechanisms within MLLMs. As shown in Figure \ref{attn_js}, we examine the Jensen-Shannon (JS) divergence between the attention of different decode layers across various inputs, and find that the attention patterns between adjacent layers are highly similar. This phenomenon indicates that a significant portion of the attention layers are essentially redundant and the ratio of these redundant layers exceeds 50\%! Inspired by the shared experts concept in Mixture of Experts (MOE) \cite{mixtral}, 
we aim to group the redundant attention layers into sparse shared attention blocks, where the redundant cost can be reused. By allocating shared attention to decode layers with similar attention patterns, we reduce attention computations and the storage for KV cache.

Based on the above insights, we propose a novel \textit{layer-wise} inference acceleration framework in this paper. Specifically, we decompose the decoding process into multiple \textbf{L}azy \textbf{B}lock (\textbf{LB}), where adjacent layers exhibit highly similar attention patterns. These layers are equipped with Lazy Attention (\textbf{LA}) to inherit computation results from previous layers, thereby reducing redundancy in the current attention calculations. Furthermore, we introduce \textbf{Q Cache}, a novel dynamic caching mechanism that enables the reuse of queries in \textbf{LA}. Due to the distribution gap between text and visual modalities, we propose two distinct granularity modes for LA: Global Lazy Attention (\textbf{GLA}) and Visual Lazy Attention (\textbf{VLA}) that are applied to all tokens and visual tokens, respectively. Based on LA and Q cache, layers within a LB share a portion of the K cache, effectively reducing the KV cache memory budget. Additionally, our \textit{layer-wise} approach is orthogonal to existing \textit{token-wise} methods, allowing for seamless integration of both techniques in scenarios with extreme resource constraints. Extensive experiments demonstrate that our method significantly reduces the model's FLOPs, parameters, KV cache memory footprint and latency, enhancing the throughput with minimal performance degradation. In summary, our main contributions can be outlined as follows:
\begin{itemize}
\item We perform an in-depth investigation of attention mechanisms in MLLMs, uncovering that the attention patterns between adjacent layers are highly similar and attention in over 50\% layers is invalid.
Based on this insight, we present a novel \textit{layer-wise} acceleration framework to remove the potential redundancy, which is orthogonal to the existing \textit{token-wise} optimization scheme.
\item We propose Lazy Attention, a novel and efficient attention mechanism that encourages the model to share the attention across adjacent layers. 
\item We introduce Q Cache a novel dynamic cache that allows queries to be reused in Lazy Attention. It is a low-overhead caching mechanism that is fully compatible with both KV cache and Flash Attention.
\item We validate the effectiveness of our method over existing methods and quantify the impact of specific components across a wide range of multimodal datasets. It can simultaneously reduces KV cache memory, FLOPs and latency of MLLMs, while achieves a 1.5$\times$ increase in throughput with negligible impact on accuracy.
\end{itemize}

%% file: sec/related.tex
\section{Related Work}
\label{sec:related}

%-------------------------------------------------------------------------
\textbf{Projector-based Token Compression.} Due to the cumbersome nature of visual tokens, several prior methods have endeavored to compress visual information by devising a variety of intricate projectors situated between the vision encoder and LLM component\cite{llava1.5,llava-next,visual_anchor,video-llava}. Cross-attention based method such as Resampler~\cite{flamingo}, Q-former~\cite{blip2} and Video-LLaMA \cite{video-llama} introduce a few learnable tokens to represent visual content. 
Convolutions combined method Honeybee~\cite{honeybee} leverage traditional convolutional operations to aggregate visual features.
Naive downsample method 
DeCo~\cite{Deco} uses 2D adaptive average pooling to merge visual patch. Tokenpacker~\cite{tokenpacker} and {$M^3$}~\cite{m3} utilize downsampling to capture multi-granularity feature maps.
These existing methods predominantly focus on token-wise redundancy and frequently introduce additional parameters that necessitate fine-tuning. In contrast, our method specifically targets layer-wise redundancy without the introduction of any additional parameters. \\
%-------------------------------------------------------------------------

\begin{figure*}[t]
\centering
\includegraphics[width=0.93\textwidth]{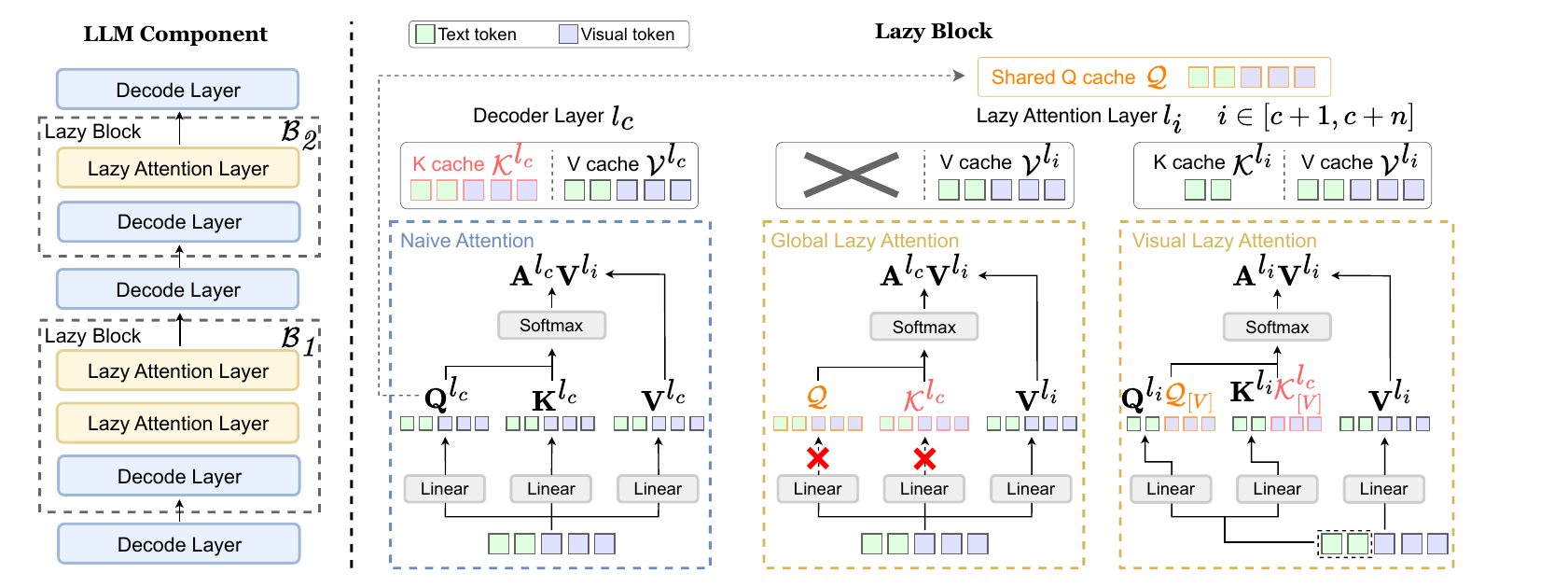}
\caption{The diagram of the entire framework. \textbf{Left}: The optimized LLM component within a MLLM. It consists of several standard decode layers and multiple custom lazy blocks. Each \textit{lazy block} contains one regular decode layer along with several lazy attention layers. \textbf{Right}: The internal structure of a lazy block. The distinctive characteristic of the lazy attention layer lies in its replacement of the naive attention with \textit{lazy attention} (Global Lazy Attention or Visual Lazy Attention), which directly utilizes the queries and keys from the first decode layer within the block. To avoid the full computation typically associated with naive attention, the lazy attention focuses on eliminating redundant computations. This design helps to optimize memory cost while managing computational complexity effectively.
}
\label{framework}
\end{figure*}
\noindent
\textbf{Visual Token Pruning.} Under the train-free paradigm, visual token pruning in multimodal models can be categorized into two routes: Before LLM \cite{purmaerge,hired,retrieval} and In LLM \cite{fastv,VTW,sparsevlm,st3}. These methods pick tokens through the attention mechanism within transformer~\cite{attention,ViT} and then prune or merge the non-critical tokens \cite{pyramiddrop,vl-cache,crossget,tome}. 
For the former route, LLaVA-Prumerge~\cite{purmaerge} and HiRED \cite{hired} adaptively prune visual tokens in vision encoder \cite{ViT,CLIP}, but ignoring the modal interactions between text and images. The latter category (In LLM) aims to evict visual tokens in decode layers of LLM component. FastV \cite{fastv} and VTW \cite{VTW} prune visual tokens once at shallow and deep layers, respectively. SparseVLM \cite{sparsevlm}, ST$^3$ \cite{st3} and PyramidDrop \cite{pyramiddrop} progressively discards visual tokens as the layers deepen. Furthermore, VL-Cache \cite{vl-cache} find that dynamic pruning ratio at each layer is more reasonable. 
These methods often lack direct compatibility with flash-attention \cite{flashattention,flashattention2} due to their dependence on attention maps. Similar to projector-based approaches, their insights into redundancy in attention computations are primarily confined to a token-wise perspective.

%% file: sec/method.tex
\begin{figure*}[t]
\centering
\includegraphics[width=0.91\textwidth]{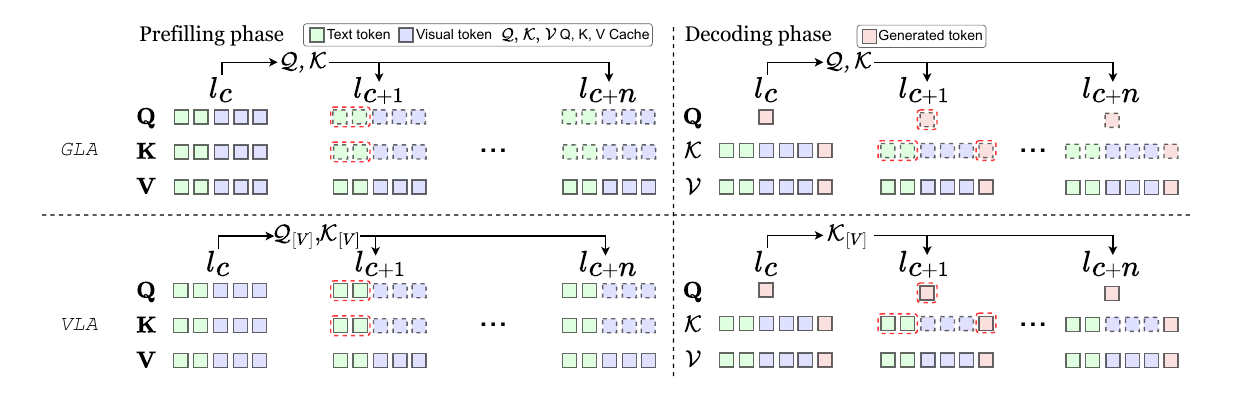}
\caption{A comprehensive visualization of the Q, K, V and caches in each layer of the lazy block in the prefilling and decoding stage under two modes. Tokens bounded by dashed lines indicate that they are not computed by the current layer or are not retained in the KV cache of this layer. The red dashed box highlights the main differences between GLA and VLA.
}
\label{flow}
\end{figure*}
\section{Method}
\label{sec:method}
% In this section, we begin with a concise introduction on attention mechanism and a overview of our framework. We then delve into how to obtain the Lazy Block and implementation details of Lazy attention and Q Cache, followed by a theoretical analysis of budget and efficiency. 
\subsection{Preliminary}
% \subsubsection{Review on Attention}
\noindent \textbf{Review on Attention.}
The inference of MLLMs is primarily divided into two phases: the prefilling phase and the decoding phase.
% We first analyze the causal attention within the LLM component of MLLM in these two stages. 
Precisely, the self-attention computation at $l$-th decode layer in prefilling can be formulated by:
\begin{equation}\label{eq3}
\begin{split}
    \mathbf{Q}^l = \mathbf{X}^l\mathbf{W}^l_\mathbf{Q};\quad \mathbf{K}^l = \mathbf{X}^l\mathbf{W}^l_\mathbf{K};\quad
    \mathbf{V}^l = \mathbf{X}^l\mathbf{W}^l_\mathbf{V}, \\
% \end{equation}
% \begin{equation}
    \mathbf{A}^l = \text{Softmax} \left( \frac{\mathbf{Q}^l (\mathbf{K}^l)^\mathsf{T}+\mathbf{M}}{\sqrt{D}} \right); \quad\mathbf{O}^l = \mathbf{A}^l\mathbf{V}^l,
\end{split}
\end{equation}
where $\mathbf{X}^l$ refers to the input of the $l$-th layer, including text and visual tokens; $\mathbf{W}^l_\mathbf{Q}$, $\mathbf{W}^l_\mathbf{K}$ and $\mathbf{W}^l_\mathbf{V}$ are linear projector for query, key and value; $\mathbf{A}^l$ denotes a lower triangular attention matrix and $\mathbf{O}^l$ means the output of this layer. In this phase, $\mathbf{K}^l$ and $\mathbf{V}^l$ will be stored in the KV cache.

In the decoding phase, the previous keys and values will be loaded from KV cache for self-attention calculations: 
\begin{equation}\label{eq4}
\begin{split}
    \mathbf{Q}^l = \mathbf{x}^l\mathbf{W}^l_\mathbf{Q};
\mathbf{K}^l = [\mathcal{K}^l, \mathbf{x}^l\mathbf{W}^l_\mathbf{K}];
    \mathbf{V}^l = [\mathcal{V}^l, \mathbf{x}^l\mathbf{W}^l_\mathbf{V}], \\
    \mathbf{A}^l = \text{Softmax} \left( \frac{\mathbf{Q}^l (\mathbf{K}^l)^\mathsf{T}}{\sqrt{D}} \right); \quad\mathbf{O}^l = \mathbf{A}^l\mathbf{V}^l,\quad
\end{split}
\end{equation}
where $\mathbf{x}^l$ denotes the generated text token as the input token; $\mathcal{K}^l$, $\mathcal{V}^l$ are KV cache for $l$-th layer; $[\cdot,\cdot]$ represents concatenation operation. 
% For clarity, here illustrates single-head self-attention. In the subsequent implementation, we average across all attention heads.
For clarity, the following illustration describes single-head self-attention. In the subsequent implementation, we average results across all attention heads.

% \subsubsection{Overview of Framework}
\noindent \textbf{Overview.}
Figure \ref{framework} illustrates the overview of our method. We first aggregate decode layers with high attention similarity into a single lazy block upon the attention distributions between adjacent layers. Each lazy block consists of a conventional decode layer and several lazy attention layers. Unlike naive attention, lazy attention avoids redundant attention computations by utilizing shared Q and K cache. The right shows two modes of lazy attention, while in implementation, the model will be optimized by only one mode.
\subsection{Lazy Attention}
To aggregate the adjacent decode layers with high attention similarity into a lazy block, we first need to empirically analyze the attention distributions across all layers. However, extracting the full attention matrix for analysis is expensive. Thus, we utilize the last row $\mathbf{a}^l$ of the attention matrix $\mathbf{A}^l$ to compute the layer-wise similarity of attention, as the last token can capture the attention scores with all previous tokens \cite{fastv, tova}. Specifically, we examine the similarity of layer-wise attention across $N$ input samples. The attention similarity between any two distinct layers $l_m$ and $l_n$ can be quantified by the Jensen-Shannon (JS) divergence \cite{jensen}: 
\begin{equation}
\begin{split}
    \mathbf{S}(l_m, l_n) = \frac{1}{N}\sum_i^N{\text{JS}(\mathbf{a}_i^{l_m}, \mathbf{a}_i^{l_n})},\quad\quad\quad \\
% \end{equation}
% \begin{equation}
    \text{JS}(\mathbf{a}_i^{l_m}, \mathbf{a}_i^{l_n})= \frac{1}{2}\bigg[\text{KL}(\mathbf{a}_i^{l_m}||\mathbf{m})+\text{KL}(\mathbf{a}_i^{l_n}||\mathbf{m})\bigg],
\end{split}
\end{equation}
where $\mathbf{m}=(\mathbf{a}_i^{l_m}+\mathbf{a}_i^{l_n})/2$ is the mean distribution, and $\text{KL}(\cdot||\cdot)$ denotes the Kullback-Leibler (KL) divergence.
We adopt the Jensen-Shannon divergence for its symmetry and boundedness properties, which ensure a more stable and reliable pattern determination. 

Empirical analysis indicates that the adjacent decode layers in MLLMs exhibit high attention similarity as shown in Figure \ref{attn_js}. This intriguing phenomenon inspires us to explore new approaches in model optimization from a layer-wise perspective. We intuitively believe that these adjacent decode layers can share all or part of attention pattern. 

After obtaining the attention similarity between layers, we aggregate adjacent layers with high similarity into ``Lazy Block'' $\mathcal{B}\in\{l_c, l_{c+1}, ..., l_{c+n}\}$, as the following formula:
\begin{equation}
    \mathbf{S}(l_i, l_{i+1}) < \epsilon\quad \forall{l_i}\in\mathcal{B},
\end{equation}
where $0<\epsilon \leq 1$ is a hyper-parameter. Here we set a threshold to limit the size of blocks $\mathcal{B}$, ensuring that attention is shared only among every small ranges of adjacent layers. This approach prevents the accumulating errors introduced by the diminishing attention similarity between distant layers. We conduct experiments on several MLLMs and find that the proportion of lazy attention layers in their LLM decoder layers exceeds 50\%. The attention mechanism within the lazy attention layers is termed as ``Lazy Attention''. We categorize it into two distinct modes Global Lazy Attention (\textbf{GLA}) and Visual Lazy Attention (\textbf{VLA}): \textbf{GLA} shares all attention pattern while \textbf{VLA} shares visual pattern.

We tailor a sharing strategy for lazy attention within each block: except for the first decode layer $l_c$, subsequent lazy attention share its attention patterns. Given the attention matrix is computed by $\textbf{Q}$ and $\textbf{K}$, we directly implement sharing strategy on them. The benefit is its seamless compatibility with existing inference frameworks, including KV cache and flash attention. We then define the Q Cache $\mathcal{Q}=\mathbf{Q}^{l_c}$, a dynamic shared cache that updates as the block changes. 
\subsection{Q Cache}
\label{lazy attention}
In this section, we discuss the usage of Q Cache in GLA and VLA modes, and its integration with KV Cache. Following formula outlines the complete pipeline of Q Cache within lazy attention for $l_i$-th layer in a block ($i\in[c+1, c+n]$. We decompose this process into prefilling and decoding stages as the query is a single text token and does not include visual tokens during the decoding phase): \\
\textbf{\textit{GLA (Prefilling)}}:
\begin{equation}
    \mathbf{Q}^{l_i} = \mathcal{Q};\quad \mathbf{K}^{l_i} = \mathcal{K}^{l_c};\quad \mathbf{V}^{l_i} = \mathbf{X}^{l_i}\mathbf{W}_\mathbf{V}^{l_i}
\end{equation}
\textbf{\textit{GLA (Decoding)}}:
\begin{equation}
    \mathbf{Q}^{l_i} = \mathcal{Q};\quad \mathbf{K}^{l_i} = \mathcal{K}^{l_c};\quad \mathbf{V}^{l_i} = [\mathcal{V}^{l_i}, \mathbf{x}^{l_i}\mathbf{W}_\mathbf{V}^{l_i}]
\end{equation}
Under the GLA mode, all layers within the block share the Q cache $\mathcal{Q}$ and K cache $\mathcal{K}^{l_c}$ corresponding to all tokens, while value $\mathbf{V}$ will be computed as normal at each layer. GLA completely eliminates the computation of the query and key mappings, as well as the storage of all K caches in lazy attention layers within a block. \\
\textbf{\textit{VLA (Prefilling)}}:
\begin{equation}
\begin{split}
    \mathbf{Q}^{l_i} = \bigg[\mathbf{X}^{l_i}_{[T]}\mathbf{W}^{l_i}_\mathbf{Q}, \mathcal{Q}_{[V]}\bigg];~
    \mathbf{K}^{l_i} = \bigg[\mathbf{X}^{l_i}_{[T]}\mathbf{W}^{l_i}_\mathbf{K}, \mathcal{K}^{l_c}_{[V]}\bigg]\\
    \mathbf{V}^{l_i} = \mathbf{X}^{l_i}\mathbf{W}_\mathbf{V}^{l_i},\quad\quad\quad\quad\quad\quad\quad
\end{split}
\end{equation}
\textbf{\textit{VLA (Decoding)}}:
\begin{equation}
\begin{split}
    \mathbf{Q}^{l_i} = \mathbf{x}^{l_i}\mathbf{W}^{l_i}_\mathbf{Q};\quad
    \mathbf{K}^{l_i} = \bigg[\mathcal{K}^{l_i}_{[T]}, \mathcal{K}^{l_c}_{[V]}, \mathbf{x}^{l_i}\mathbf{W}^{l_i}_\mathbf{K}\bigg]\quad\\
    \mathbf{V}^{l_i} = \bigg[\mathcal{V}^{l_i}, \mathbf{x}^{l_i}\mathbf{W}_\mathbf{V}^{l_i}\bigg],\quad\quad\quad\quad\quad
\end{split}
\end{equation}
where $\mathbf{X}^{l_i}_{[T]}$ means all text tokens in the input sequence of layer $l_i$. Here we simplify the actual positions of text and visuals tokens by representing them through concatenation. $\mathcal{Q}_{[V]}$ is visual part in Q cache; $\mathcal{K}_{[T]}$, $\mathcal{K}_{[V]}$ represent the text and visual cache in K cache. Different from GLA, VLA is a more fine-grained mode that retains the mapping operations of text tokens in query and key while sharing the visual parts of the Q cache and K cache. The rationale behind this design is that the number of visual tokens is often much larger than that of text tokens in MLLMs. Therefore, sharing attention only for visual tokens can yield significant cost savings. 
We implement both GLA and VLA and present a comprehensive analysis on their effectiveness.

\begin{table*}[t]
%\scriptsize
\footnotesize
%\fontsize{9}{9}
% \footnotesize
\renewcommand{\arraystretch}{0.6}
\setlength{\tabcolsep}{0.8mm}
\centering
\begin{tabular}{l|cccccccccccc}
\toprule
Methods & GQA$\uparrow$ & VQA$\uparrow$ & TextVQA$\uparrow$ & VizWiz$\uparrow$ & AI2D$\uparrow$ & SQA$\uparrow$ & MMMU$\uparrow$ & MMB$\uparrow$ & POPE$\uparrow$ & COCO$\uparrow$ & Nocaps$\uparrow$ & Flickr30K$\uparrow$ \\ 
% \midrule
% \multicolumn{13}{c}{\textit{LLaVA-v1.5-7B}} \\
\midrule
\textit{LLaVA-v1.5-7B} & 61.94 & 76.64 & 46.10 & 54.38 & 55.25 & 69.51 & 35.30 & 64.09 & 85.13 & 110.43 & 105.53 & 74.89 \\
% FastV$\ddagger$~\cite{fastv} & 9.38T & 70.80ms & 1862.12 & 55.34 & 68.77 & 35.00 & 63.83 & 85.13 \\
FastV (ECCV24) & 60.32 & 75.86 & 45.52 & 54.47 & 55.24 & 68.72 & 35.00 & 63.81 & 84.97 & 110.48 & 105.31 & \textbf{74.67} \\
VTW (AAAI25) & 55.14 & 66.28 & 16.08 & 51.00 & \textbf{55.30} & 69.30 & 36.10 & 63.92 & 85.10 & 67.22 & 95.78 & 40.69 \\
HiRED (AAAI25) & 59.42 & 75.54 & 45.60 & 53.55 & 52.85 & 68.12 & 36.10 & 63.57 & 84.99 & 102.27 & 100.10 & 72.86 \\
PruMerge+ (ICCV25) & 57.39 & 72.77 & 39.55 & 54.50 & 54.18 & 69.36 & 36.30 & 61.43 & 84.09 & 102.96 & 99.69 & 70.17 \\
SparseVLM (ICML25) & 57.24 & 73.21 & 42.65 & 54.39 & 54.92 & 69.01 & 35.10 & 64.00 & 84.88 & 103.69 & 98.88 & 69.89 \\
\midrule
\rowcolor{gray!16}
GLA (Ours) & \textbf{61.96} & 76.06 & 42.21 & 53.43 & 52.89 & 66.46 & 30.30 & 60.62 & 84.02 & 105.75 & 100.35 & 70.74 \\
\rowcolor{gray!16}
VLA (Ours) & 61.93 & \textbf{76.38} & \textbf{45.70} & \textbf{54.58} & 54.98 & \textbf{69.43} & \textbf{36.80} & \textbf{64.88} & \textbf{85.13} & \textbf{110.51} & \textbf{105.60} & 74.36 \\
\midrule
\midrule
\textit{LLaVA-v1.5-13B} & 63.28 & 78.19 & 48.69 & 56.61 & 59.26 & 72.78 & 34.90 & 68.81 & 85.66 & 115.55 & 109.31 & 79.56 \\
% FastV$\ddagger$~\cite{fastv} & 17.81T & 128.49ms & 1817.55 & 58.87 & 72.98 & 33.90 & 68.30 & 86.17 \\
FastV (ECCV24) & 62.62 & 77.66 & \textbf{48.32} & 56.47 & 58.84 & 72.88 & 33.90 & 68.30 & 85.26 & \textbf{115.51} & \textbf{108.83} & 79.48\\
VTW (AAAI25) & 60.61 & 75.34 & 34.25 & 55.11 & \textbf{59.22} & 72.68 & 34.80 & 68.41 & 85.71 & 101.97 & 96.31 & 65.80 \\
HiRED (AAAI25) & 59.69 & 76.63 & 47.92 & 55.10 & 55.73 & 72.34 & \textbf{34.90} & 67.18 & 85.58 & 107.80 & 104.46 & 78.00 \\
PruMerge+ (ICCV25) & 58.16 & 73.89 & 41.57 & 55.81 & 57.93 & \textbf{73.07} & 34.80 & 65.21 & 85.32 & 105.11 & 101.05 & 71.13 \\
SparseVLM (ICML25) & 58.69 & 74.92 & 45.32 & 56.46 & 57.16 & 72.77 & 
\textbf{34.90} &  67.53 & 85.48 & 109.52 & 103.97 & 75.33 \\
\midrule
\rowcolor{gray!16}
GLA (Ours) & 62.98 & 77.77 & 44.14 & 55.90 & 56.42 & 70.06 & 31.70 & 65.15 & 84.53 & 110.84 & 103.67 & 76.10 \\
\rowcolor{gray!16}
VLA (Ours) & \textbf{63.31} & \textbf{78.03} & 47.83 & \textbf{57.42} & 59.19 & 72.55 & \textbf{34.90} & \textbf{68.61} & \textbf{85.73} & 115.44 & 108.53 & \textbf{79.54} \\
\midrule
\midrule
\textit{LLaVA-NEXT-7B} & 64.24 & 79.89 & 64.86 & 60.66 & 65.25 & 70.15 & 35.40 & 67.10 & 86.40 & 99.87 & 88.37 & 68.47 \\
% FastV$\ddagger$~\cite{fastv} & 30.95T & 408.61ms & 1806.93 & 65.12 & 69.21 & 35.70 & 66.41 & \textbf{87.78}\\
FastV (ECCV24) & 63.87 & 79.63 & \textbf{63.79} & 60.19 & 65.02 & 69.11 & 35.10 & 66.31 & 85.98 & 98.54 & \textbf{86.48} & 67.18 \\
VTW (AAAI25) & 55.84 & 68.45 & 18.88 & 55.07 & \textbf{65.21} & \textbf{69.91} & 35.20 & 66.92 & 86.21 & 82.50 & 43.76 & 57.81 \\
HiRED (AAAI25) & 61.82 & 78.51 & 63.65 & 59.71 & 62.77 & 68.74 & 35.40 & 66.96 & 86.01 & 93.79 & 84.28 & 65.41 \\
PruMerge+ (ICCV25) & 60.45 & 76.73 & 57.65 & 60.35 & 63.80 & 69.00 & 35.50 & 64.62 & 85.87 & 94.11 & 83.72 & 64.29\\
SparseVLM (ICML25) & 59.86 & 76.50 & 60.00 & \textbf{60.40} & \textbf{65.21} & 69.78 & 35.10 & 66.44 & 85.53 & 94.71 & 82.80 & 63.22 \\
\midrule
\rowcolor{gray!16}
GLA (Ours) & 63.55 & 79.27 & 60.11 & 58.79 & 62.23 & 66.40 & 32.30 &  64.56 & 85.00 & 94.48 & 81.99 & 63.81\\
\rowcolor{gray!16}
VLA (Ours) & \textbf{64.12} & \textbf{79.65} & 63.66 & 60.24 & 64.37 & 69.55 & \textbf{35.80} & \textbf{67.02} & \textbf{86.30} & \textbf{99.17} & 86.36 & \textbf{67.42} \\
\midrule
\midrule
\textit{LLaVA-NEXT-13B} & 65.43 & 80.93 & 67.12 & 63.56 & 70.27 & 73.57 & 36.10 & 69.33 & 86.10 & 101.93 & 88.18 & 66.70 \\
% FastV$\ddagger$~\cite{fastv} & 57.66T & 691.02ms & 1887.00 & 69.95 & 72.98 & 35.90 & 68.81 & 87.48 \\
FastV (ECCV24) & 64.79 & 80.72 & \textbf{66.00} & 63.25 & 69.95 & \textbf{72.98} & 35.90 & 68.47 & 85.90 & 101.29 & \textbf{87.86} & 66.13\\
VTW (AAAI25) & 61.47 & 77.01 & 43.27 & 58.50 & \textbf{70.12} & 72.47 & 35.30 & 68.94 & 86.01 & 86.86 & 43.26 & 56.52 \\
HiRED (AAAI25) & 60.72 & 78.22 & 65.96 & 62.76 & 67.08 & 72.16 & 35.90 & 68.04 & 85.83 & 96.09 & 85.27 & 64.41 \\
PruMerge+ (ICCV25) & 59.66 & 75.58 & 59.30 & 62.66 & 68.70 & 73.41 & 35.80 & 65.80 & 85.55 & 93.72 & 81.52 & 59.93 \\
SparseVLM (ICML25) & 60.01 & 76.99 & 63.72 & 63.40 & 68.09 & 73.23 & 36.00 & 68.51 & 85.77 & 97.61 & 84.87 & 62.18 \\
\midrule
\rowcolor{gray!16}
GLA (Ours) & 64.61 & 79.30 & 62.49 & 62.56 & 67.24 & 71.11 & 34.80 & 65.80 & 85.03 & 96.83 & 83.25 & 63.20\\
\rowcolor{gray!16}
VLA (Ours) & \textbf{65.33} & \textbf{80.85} & 65.26 & \textbf{63.51} & 69.17 & 72.36 & \textbf{36.10} & \textbf{69.01} & \textbf{86.10} & \textbf{101.91} & 87.77 & \textbf{66.29} \\
\bottomrule
\end{tabular}
\caption{Comparison of existing token-wise optimizing method on MLLMs on various datasets. SQA means the ScienceQA-IMG \cite{SQA} subset. MMB denotes the english subset of MMBench \cite{mmbench}. VQA, TextVQA, VizWiz are validation subset of VQA$^{\text{v2}}$ \cite{vqa}, TexTVQA \cite{textvqa}, VizWiz \cite{vizwiz}. COCO and Nocaps represent validation split in the datastes COCO2017 caption \cite{coco} and Nocaps \cite{nocaps}. Flickr30K is the test split in its original dataset \cite{flickr30k}. GQA \cite{gqa}, AI2D \cite{ai2d}, MMMU \cite{mmmu} and POPE \cite{pope} are full dataset. The best results are in \textbf{bold}.}
\label{table: main}
\end{table*}

\subsection{Efficiency Analysis}\label{cost}
\textbf{KV Cache.} Our method is compatible with KV cache \cite{kvcache} and can save a significant KV cache memory. Given layers $l_{c+1}\sim l_{c+n}$ load $\mathbf{K}^{l_i}$ from $\mathcal{K}^{l_c}$, these layers do not need to store the complete KV cache. This rule is valid regardless of whether GLA or VLA is used. Let's define the entire LLM component includes $N$ decoder layers and $n$ lazy attention layers. For GLA, the overall savings rate of the KV cache is $\frac{n}{2N}$; as for VLA, the savings rate is $\frac{|V|}{|V|+|T|}\cdot\frac{n}{2N}$ since the number of visual tokens is significantly greater than that of text ($|V|\gg|T|$), text cache retained in VLA can be considered negligible and $\frac{n}{2N}\approx\frac{|V|}{|V|+|T|}\cdot\frac{n}{2N}$. Here, we neglect the memory overhead of the Q cache as its proportion is only $\frac{1}{2N}$. Figure \ref{flow} illustrates the KV cache flow in inference for a more intuitive understanding. 
% \newline
% \textbf{Parameters} Compared to token-wise optimization methods, our method has the unique advantage of saving partial model parameters. This benefit mainly applies in GLA, where layers $l_{c+1}\sim l_{c+n}$ no longer involve $\mathbf{W}_\mathbf{Q}$ and $\mathbf{W}_\mathbf{K}$ in their attention calculations, allowing those parameters to be directly discarded. If the parameters of a linear projector account for $\alpha\%$ of the total model parameters, then GLA can save approximately $2n\cdot\alpha\%$ of the GPU memory occupied by the model.
\newline
\textbf{FLOPs and Latency.} Both GLA and VLA can reduce FLOPs and latency, as some linear computations in the attention mechanism are omitted. If the FLOPs of a linear projector in the attention mechanism accounts for $\beta\%$ of the total FLOPS of the model, then GLA and VLA can save approximately $2n\cdot\beta\%$ and $\frac{|V|}{|V|+|T|}\cdot2n\cdot\beta\%$ of the computational cost, respectively. Due to $|V|\gg|T|$, 
these two quantities are approximately equal, i.e.,
$2n\cdot\beta\%\approx\frac{|V|}{|V|+|T|}\cdot2n\cdot\beta\%$. 
% For specific benefits, please refer to the experiments in Sec. \ref{efficiency}. 
\newline
% \textbf{Flash Attention} The advantage over token-wise methods is that our technique is fully compatible with flash attention \cite{flashattention,flashattention2}. 
% Existing token pruning methods rely on the attention matrix, making it challenging to achieve direct compatibility with flash attention without additional tricks. 
% Differently, our method does not require online retrieval of the attention matrix during inference, which is not supported by flash attention. 
% However, existing token pruning methods rely on the attention matrix, making it challenging to achieve direct compatibility with flash attention without additional tricks. 

%% file: sec/experiments.tex
\setlength{\tabcolsep}{4pt}
\begin{table}[t]
% \scriptsize
%\footnotesize
%\small
%\fontsize{9}{9}
\renewcommand{\arraystretch}{0.6}
\setlength{\tabcolsep}{0.5mm}
\centering
\footnotesize
\begin{tabular}{l|cccccc}
\toprule
Method & TextVQA$\uparrow$ & AI2D$\uparrow$ & SQA$\uparrow$ & MMMU$\uparrow$ & MMB$\uparrow$ & COCO$\uparrow$ \\ 
\midrule
% \multicolumn{7}{c}{\textit{LLaVA-v1.5-7B}} \\
% \midrule
VLA & 45.70 & 54.98 & 69.43 & 36.80 & 64.88 & 110.51  \\
with FastV & 44.13 & 53.72 & 69.56 & 35.60 & 65.17 & 109.83 \\
% \midrule
% \multicolumn{7}{c}{\textit{LLaVA-v1.5-13B}} \\
% \midrule
% VLA & 47.83 & 59.19 & 72.55 & 34.90 & 67.61 & 114.44 \\
% with FastV & 47.44 & 57.85 & 71.95 & 34.60 & 67.21 & 113.30\\
\bottomrule
\end{tabular}
\caption{Examination on combination of our method and token-wise pruning method FastV with LLaVA-v1.5-7B.}
\label{table: compatibility}
\end{table}

\section{Experiments}
\label{sec:experiments}
\subsection{Main Results}
\textbf{Comparison on other methods.} In Table \ref{table: main}, we present a comparison of our method with existing MLLM acceleration optimization methods across distinct sizes and versions of the LLaVA. Our method VLA, maintains relatively stable performance across almost all datasets compared to the vanilla model and outperforms existing token-wise methods in most datasets. Notably, on fine-grained understanding datasets (GQA, TextVQA, COCO, NoCaps, Flickr30K), most token-wise methods experience significant performance degradation, especially in image captioning tasks (COCO, NoCaps, Flickr30K). This result indicates that existing token pruning methods may be unreasonable to use FLOPs and token budgets as efficiency metrics in the prefilling phase, which compromises the integrity of the KV cache and leads to failures in long text generation tasks, e.g., image captioning. 
In contrast, our layer-wise method is token-preserving. It reduces inter-layer attention redundancy by introducing the Q cache, allowing the KV cache to be reused across similar layers. This approach cleverly avoids the issue of context loss. Another intriguing phenomenon is that GLA method consistently lags behind VLA across almost all evaluations. Intuitively, we believe this is because the distribution gap between visual attention and text attention. 
We provide a detailed analysis of the underlying reasons for this result in \textit{Section Analysis}. 
In summary, our VLA can significantly reduce the KV cache footprint of the model with minimal performance degradation.
Unless otherwise stated, the following experimental sections are conducted in VLA. \\
\noindent
\textbf{Compatibility.} A major advantage of our method lies in its focus on layer-wise attention redundancy, making it compatible with existing token-wise optimizing approaches (two methods can be combined effectively). Table \ref{table: compatibility} presents the results on the combination of VLA and FastV \cite{fastv}. Upon the VLA, FastV only brings a 1\%-2\% loss in varied performance metrics, which is consistent with observations from token pruning on LLaVA in Table \ref{table: main}. This indicates that our technique is orthogonal to the token-wise optimizing route. The model optimized by Q cache and lazy attention can serve as a foundation, replacing the original MLLM, and support pruning or quantization techniques. 
% Moreover, overlaying FastV \cite{fastv} on our method can further reduce resource budget and enhance throughput (see Sec. \ref{efficiency}). 

\begin{figure}[t]
\centering
\includegraphics[width=0.47\textwidth]{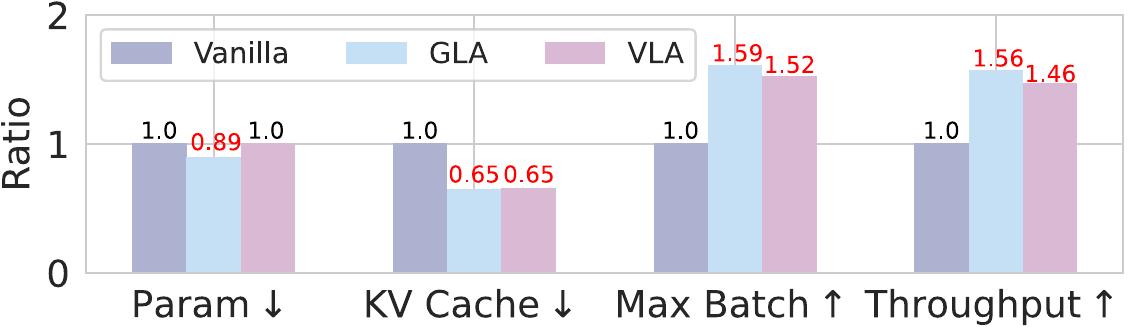}
\caption{The benefits of GLA and VLA over the Vanilla LLaVA-NEXT-7B model in parameters, KV cache, max batchsize and throughput in decoding phase.
}
\label{fig: efficiency}
\end{figure}

\noindent
\textbf{Efficiency Evaluation.} As shown in Table \ref{table: efficiency}, we measure the model's FLOPs and latency in the prefilling phase, as well as the KV cache footprint under batchsize=1. Except the parameters metric of VLA, other metrics demonstrate significant improvements compared to the Vanilla model. 
% This result aligns with our theoretical analysis in the Sec. \ref{cost}. 
Figure \ref{fig: efficiency} demonstrates the benefits of GLA and VLA over LLaVA-NEXT-7B on a single 80GB A100 GPU. With an input length of 2048, the maximum batch size increases from 21 to 32, due to the reduction in KV cache memory. Overall, \textbf{GLA and VLA save 35\% KV cache memory, while achieving 1.6$\times$ and 1.5$\times$ throughput}, respectively.

\subsection{Sanity Check}
\label{Sanity Check}
\begin{figure}[t]
\centering
\includegraphics[width=0.47\textwidth]{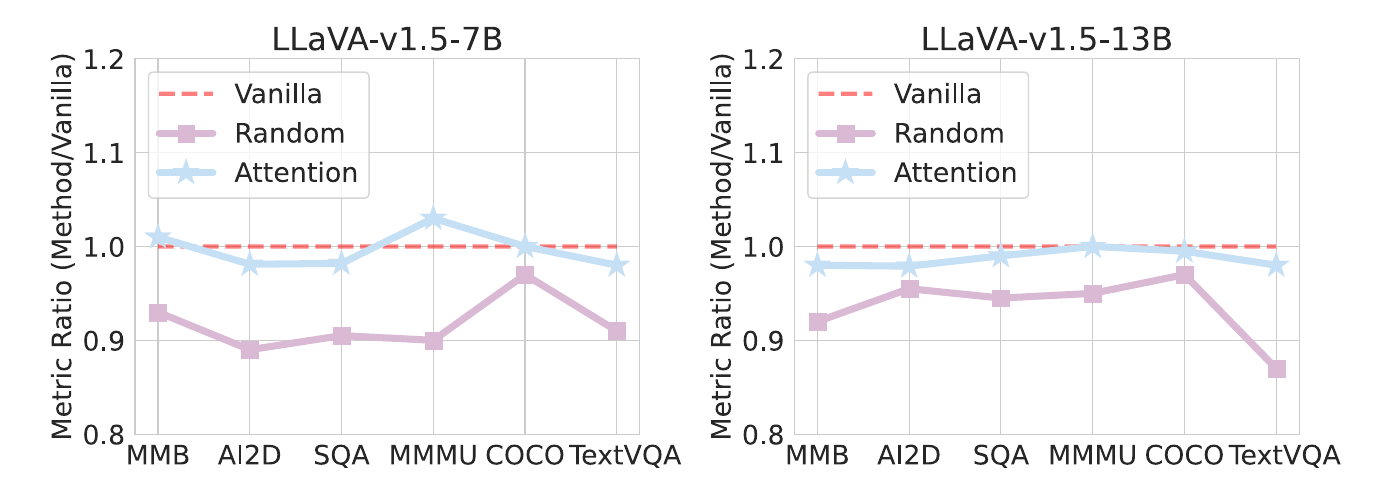}
\caption{Comparison of random selection and attention similarity metric on partial datasets. 
}
\label{random}
\end{figure}

\textbf{Random Layers Fail in Performance}. We conduct a validity test on our method by constructing lazy blocks via attention similarity. Specifically, we construct lazy blocks by grouping adjacent decode layers with random selection. For a fair comparison, we keep the number of lazy blocks and lazy layers consistent across both methods. Utilizing the attention similarity metric to construct lazy blocks, significantly outperformed the random selection method on both LLaVA-v1.5-7B/13B in 6 datasets. As shown in Figure \ref{random}, the experiment demonstrates that the high similarity of attention is the primary factor contributing to the model's performance retention by the VLA.

\setlength{\tabcolsep}{4pt}
\begin{table}[t]
%\scriptsize
%\footnotesize
%\small
%\fontsize{9}{9}
\renewcommand{\arraystretch}{0.6}
\setlength{\tabcolsep}{0.32mm}
\centering
\footnotesize
\begin{tabular}{l|cc|ccc}
\toprule
Method  & FLOPs (T) & Latency (ms) & Params (B) & KV Cache (GB) & \\
\midrule
% \multicolumn{6}{c}{\textit{LLaVA-v1.5-7B}} \\
% \midrule
Vanilla & 30.88 & 142.60 & 6.76 & 1.01 \\
GLA & 18.17\decrease{40.9\%} & 66.79\decrease{53.2\%} & 5.98\decrease{11.5\%} & 0.35\decrease{65.0\%} \\
VLA & 18.49\decrease{40.1\%} & 67.21\decrease{52.9\%} & 6.76 & 0.36\decrease{64.0\%} \\
% \midrule
% \multicolumn{6}{c}{\textit{LLaVA-NEXT-7B}} \\
% \midrule
% Vanilla & $\downarrow$35.9\% & $\downarrow$5.8\% & $\downarrow$11.5\% & $\downarrow$10.3\% \\
% \textit{mode1} & 42.21 & 56.54 & 68.91 & 33.20  \\
% \textit{mode2} & 47.83 & 58.19 &  & 34.90 \\
\bottomrule
\end{tabular}
\caption{Efficency test on LLaVA-NEXT-7B combined with FastV in prefilling phase.}
\label{table: efficiency}
\end{table}

\begin{figure}[t]
\centering
\includegraphics[width=0.4\textwidth]{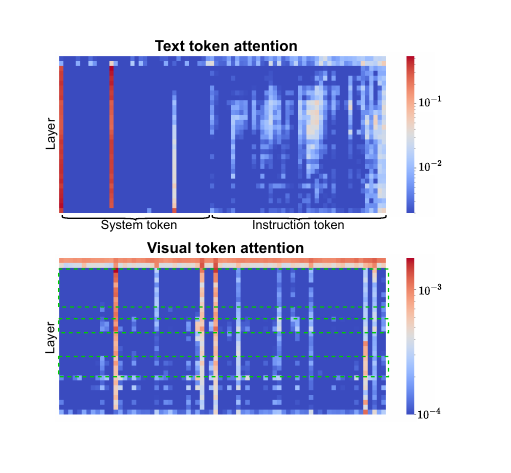}
\caption{Cross-layer visualization of text and visual tokens. For a clearer presentation, average pooling was applied to the visual tokens. The adjacent layers with a high attention similarity are highlighted with dashed bounding boxes. 
}
\label{attn_dis}
\end{figure}

% \setlength{\tabcolsep}{4pt}
% \begin{table}[t]
% %\scriptsize
% %\footnotesize
% %\small
% %\fontsize{9}{9}
% \renewcommand{\arraystretch}{0.6}
% \setlength{\tabcolsep}{0.5mm}
% \centering
% \footnotesize
% \begin{tabular}{l|cccccc}
% \toprule
% Method & TextVQA$\uparrow$ & AI2D$\uparrow$ & SQA$\uparrow$ & MMMU$\uparrow$ & MMB$\uparrow$ \\ 
% \midrule
% % \multicolumn{7}{c}{\textit{InternVL2-8B}} \\
% % \midrule
% \textbf{InternVL2-8B} & 76.88 & 82.42 & 97.07 & 48.44 & 81.62  \\
% FastV & 73.97 & \textbf{80.25} & 94.97 & 47.68 & 79.78 \\
% \rowcolor{gray!16}
% VLA(Ours) & \textbf{74.22} & 80.10 & \textbf{95.23} & \textbf{47.85} & \textbf{80.77}  \\
% \midrule
% % \multicolumn{7}{c}{\textit{Qwen2-VL-7B}} \\
% % \midrule
% \textbf{Qwen2-VL-7B} & 82.07 & 80.02 & 95.09 & 50.56 & 79.04 \\
% FastV & \textbf{80.28} & 77.82 & 92.78 & 49.56 & 76.29 \\
% \rowcolor{gray!16}
% VLA(Ours) & 80.01 & \textbf{78.64} & \textbf{93.01} & \textbf{50.14} & \textbf{77.98} \\
% \bottomrule
% \end{tabular}
% \caption{Test on Qwen2-VL and InternVL2 with VLA.}
% \label{table: mllms}
% \end{table}

\begin{figure}[t]
\centering
\includegraphics[width=0.435\textwidth]{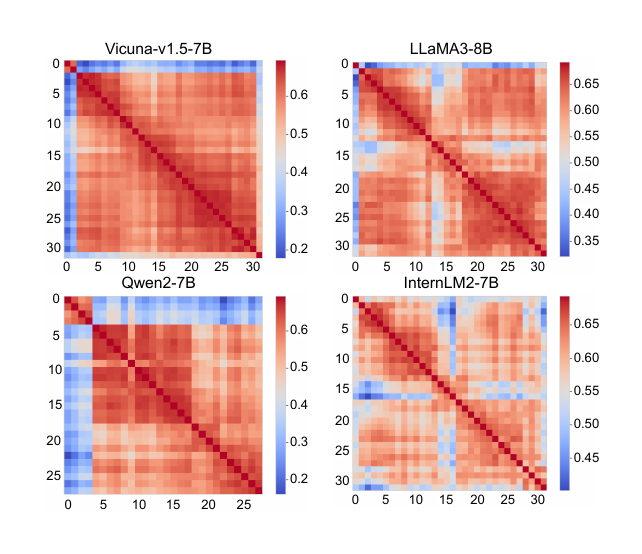}
\caption{Attention scores similarity across all layers in LLMs. A large value means high similarity. It indicates that adjacent layers have more similar attention patterns.
}
\label{attn_llm}
\end{figure}

\subsection{Analysis}
\label{analysis}
\textbf{Why VLA is Better than GLA}. To reveal the reason for that VLA performing better than GLA, we conduct a more granular analysis of attention in the decode layers. Specifically, we decompose attention into visual attention and text attention, visualizing their distributions across layers. As shown in Figure \ref{attn_dis}, our findings are as follows: \\
\textbf{Finding1}: \textit{Some text tokens that lack actual significance receive a majority of the attention scores, and these positions remain consistent across all layers}. It explains the reason for the low JS Divergence between most adjacent layers in Figure \ref{attn_js} --- the presence of these unusually high attention tokens dominates the computation results. \\
\textbf{Finding2}: \textit{The attention of meaningful text (instruction tokens) shows larger variation across layers}. Combined with Finding1, we infer that JS Divergence can be somewhat misleading, causing certain layers to incorrectly reuse text attention that have low similarity. We believe this is the underlying principle behind the poor performance of GLA. \\
\textbf{Finding3}: \textit{The attention scores on visual tokens are significantly lower than those of text tokens, while exhibiting higher inter-layer similarity}. This can explain why VLA achieves more stable performance, as VLA merely inherits visual components from the preceding layers. Since visual attention has higher inter-layer similarity and constitutes a smaller proportion of the total attention, applying lazy attention on visual tokens has a less impact on the output.

\noindent
\textbf{Attention Similarity in LLMs}. We conduct an attention analysis in the pure text on the LLMs \cite{vicuna,llama3,qwen2,internlm2} to help us understand the reasons behind the aforementioned phenomenon in MLLMs. From Figure \ref{attn_llm} (values are calculated by $\text{ln}2-\textbf{S}(l_m,l_n)$), lazy blocks are present across different LLMs with varied distribution. This suggests that inter-layer attention redundancy is a widespread occurrence. Notably, the patterns exhibited in Vicuna are almost identical to those in LLaVA (see Figure \ref{attn_js}), as the first two layers and the last layer exhibit lower similarity compared to the other layers, which further solidifies our claim that the lazy characteristics of a MLLM are inherited from its LLM.

\noindent
\textbf{Lazy Instruction Tuning}. Based on the above findings, we boldly hypothesize that we can directly implement our method on a LLM that has not undergone vision-text fine-tuning. During the second stage of fine-tuning the LLaVA model (instruction tuning phase), we introduce Q cache and lazy attention on the Vicuna model. As in Table \ref{table: LLM}, the results of fine-tuning the original Vicuna weights directly are comparable to the default method, further validating our observations regarding LLMs in the previous section. This outcome inspires us to adopt Q cache and lazy attention methods during the instruction tuning phase, thereby avoiding the overhead associated with post-training. 

\setlength{\tabcolsep}{4pt}
\begin{table}[t]
%\scriptsize
%\footnotesize
%\small
%\fontsize{9}{9}
\renewcommand{\arraystretch}{0.6}
\setlength{\tabcolsep}{0.7mm}
\centering
\footnotesize
\begin{tabular}{l|cccccc}
\toprule
LLM & TextVQA$\uparrow$ & AI2D$\uparrow$ & SQA$\uparrow$ & MMMU$\uparrow$ & MMB$\uparrow$ & COCO$\uparrow$ \\ 
\midrule
Vicuna $\dagger$ & 45.70 & 54.98 & 69.43 & 36.80 & 64.88 & 110.51 \\
Vicuna $\ddagger$ & 45.12 & 54.95 & 69.06 & 37.00 & 65.38 & 109.03 \\
\bottomrule
\end{tabular}
\caption{Comparison on weights of LLM component within LLaVA-v1.5-7B. ``$\dagger$'' means applying our method for post-training on the Vicuna within a pre-trained LLaVA. ``$\ddagger$'' denotes directly developing our method on Vicuna for LLaVA instruction tuning.}
\label{table: LLM}
\end{table}

%% file: sec/conclusion.tex
\section{Conclusion}
This paper presents a novel cross-layer shared cache and a memory-friendly attention mechanism. We first dive into attention properties within MLLMs and discover that visual attention in the majority of decode layers is redundant. Motivated by this insight, we propose Q cache and lazy attention, which significantly reduces the KV cache memory footprint without compromising accuracy. 
Furthermore, we conduct an in-depth analysis of the attention redundancy phenomenon in MLLMs, positing that this characteristic of lethargy is inherited from pre-trained LLMs. 
We hope this work can provide inspiration for designing fancy cache and instruction tuning techniques, sparking greater efforts from the research community dedicated to enhancing our understanding of intriguing phenomena exhibited by MLLMs.
\label{sec:conclusion}

%% file: main.bbl
\begin{thebibliography}{51}
\providecommand{\natexlab}[1]{#1}

\bibitem[{Agrawal et~al.(2019)Agrawal, Desai, Wang, Chen, Jain, Johnson, Batra, Parikh, Lee, and Anderson}]{nocaps}
Agrawal, H.; Desai, K.; Wang, Y.; Chen, X.; Jain, R.; Johnson, M.; Batra, D.; Parikh, D.; Lee, S.; and Anderson, P. 2019.
\newblock Nocaps: Novel object captioning at scale.
\newblock In \emph{Proceedings of the IEEE/CVF international conference on computer vision}, 8948--8957.

\bibitem[{Alayrac et~al.(2022)Alayrac, Donahue, Luc, Miech, Barr, Hasson, Lenc, Mensch, Millican, Reynolds et~al.}]{flamingo}
Alayrac, J.-B.; Donahue, J.; Luc, P.; Miech, A.; Barr, I.; Hasson, Y.; Lenc, K.; Mensch, A.; Millican, K.; Reynolds, M.; et~al. 2022.
\newblock Flamingo: a visual language model for few-shot learning.
\newblock \emph{Advances in neural information processing systems}, 35: 23716--23736.

\bibitem[{Arif et~al.(2024)Arif, Yoon, Nikolopoulos, Vandierendonck, John, and Ji}]{hired}
Arif, K. H.~I.; Yoon, J.; Nikolopoulos, D.~S.; Vandierendonck, H.; John, D.; and Ji, B. 2024.
\newblock HiRED: Attention-Guided Token Dropping for Efficient Inference of High-Resolution Vision-Language Models in Resource-Constrained Environments.
\newblock \emph{arXiv preprint arXiv:2408.10945}.

\bibitem[{Bigham et~al.(2010)Bigham, Jayant, Ji, Little, Miller, Miller, Miller, Tatarowicz, White, White et~al.}]{vizwiz}
Bigham, J.~P.; Jayant, C.; Ji, H.; Little, G.; Miller, A.; Miller, R.~C.; Miller, R.; Tatarowicz, A.; White, B.; White, S.; et~al. 2010.
\newblock Vizwiz: nearly real-time answers to visual questions.
\newblock In \emph{Proceedings of the 23nd annual ACM symposium on User interface software and technology}, 333--342.

\bibitem[{Bolya et~al.(2022)Bolya, Fu, Dai, Zhang, Feichtenhofer, and Hoffman}]{tome}
Bolya, D.; Fu, C.-Y.; Dai, X.; Zhang, P.; Feichtenhofer, C.; and Hoffman, J. 2022.
\newblock Token merging: Your vit but faster.
\newblock \emph{arXiv preprint arXiv:2210.09461}.

\bibitem[{Cai et~al.(2024)Cai, Yang, Gao, and Lee}]{m3}
Cai, M.; Yang, J.; Gao, J.; and Lee, Y.~J. 2024.
\newblock Matryoshka Multimodal Models.
\newblock \emph{arXiv preprint arXiv:2405.17430}.

\bibitem[{Cha et~al.(2024)Cha, Kang, Mun, and Roh}]{honeybee}
Cha, J.; Kang, W.; Mun, J.; and Roh, B. 2024.
\newblock Honeybee: Locality-enhanced projector for multimodal llm.
\newblock In \emph{Proceedings of the IEEE/CVF Conference on Computer Vision and Pattern Recognition}, 13817--13827.

\bibitem[{Chen et~al.(2024)Chen, Zhao, Liu, Bai, Lin, Zhou, and Chang}]{fastv}
Chen, L.; Zhao, H.; Liu, T.; Bai, S.; Lin, J.; Zhou, C.; and Chang, B. 2024.
\newblock An image is worth 1/2 tokens after layer 2: Plug-and-play inference acceleration for large vision-language models.
\newblock \emph{arXiv preprint arXiv:2403.06764}.

\bibitem[{Chen et~al.(2015)Chen, Fang, Lin, Vedantam, Gupta, Dollar, and Zitnick}]{coco}
Chen, X.; Fang, H.; Lin, T.-Y.; Vedantam, R.; Gupta, S.; Dollar, P.; and Zitnick, C.~L. 2015.
\newblock Microsoft coco captions: Data collection and evaluation server.
\newblock \emph{arXiv preprint arXiv:1504.00325}.

\bibitem[{Chiang et~al.(2023)Chiang, Li, Lin, Sheng, Wu, Zhang, Zheng, Zhuang, Zhuang, Gonzalez et~al.}]{vicuna}
Chiang, W.-L.; Li, Z.; Lin, Z.; Sheng, Y.; Wu, Z.; Zhang, H.; Zheng, L.; Zhuang, S.; Zhuang, Y.; Gonzalez, J.~E.; et~al. 2023.
\newblock Vicuna: An open-source chatbot impressing gpt-4 with 90%* chatgpt quality, March 2023.
\newblock \emph{URL https://lmsys. org/blog/2023-03-30-vicuna}, 3(5).

\bibitem[{Dai et~al.(2025)Dai, Li, Zhuang, Zhang, and Yang}]{c3vg}
Dai, M.; Li, J.; Zhuang, J.; Zhang, X.; and Yang, W. 2025.
\newblock Multi-task visual grounding with coarse-to-fine consistency constraints.
\newblock In \emph{Proceedings of the AAAI Conference on Artificial Intelligence}, volume~39, 2618--2626.

\bibitem[{Dao(2023)}]{flashattention2}
Dao, T. 2023.
\newblock Flashattention-2: Faster attention with better parallelism and work partitioning.
\newblock \emph{arXiv preprint arXiv:2307.08691}.

\bibitem[{Dao et~al.(2022)Dao, Fu, Ermon, Rudra, and Re}]{flashattention}
Dao, T.; Fu, D.; Ermon, S.; Rudra, A.; and Re, C. 2022.
\newblock Flashattention: Fast and memory-efficient exact attention with io-awareness.
\newblock \emph{Advances in Neural Information Processing Systems}, 35: 16344--16359.

\bibitem[{Dong et~al.(2024)Dong, Zhang, Zang, Cao, Wang, Ouyang, Wei, Zhang, Duan, Cao et~al.}]{internlm2}
Dong, X.; Zhang, P.; Zang, Y.; Cao, Y.; Wang, B.; Ouyang, L.; Wei, X.; Zhang, S.; Duan, H.; Cao, M.; et~al. 2024.
\newblock Internlm-xcomposer2: Mastering free-form text-image composition and comprehension in vision-language large model.
\newblock \emph{arXiv preprint arXiv:2401.16420}.

\bibitem[{Dosovitskiy et~al.(2020)Dosovitskiy, Beyer, Kolesnikov, Weissenborn, Zhai, Unterthiner, Dehghani, Minderer, Heigold, Gelly et~al.}]{ViT}
Dosovitskiy, A.; Beyer, L.; Kolesnikov, A.; Weissenborn, D.; Zhai, X.; Unterthiner, T.; Dehghani, M.; Minderer, M.; Heigold, G.; Gelly, S.; et~al. 2020.
\newblock An image is worth 16x16 words: Transformers for image recognition at scale.
\newblock \emph{arXiv preprint arXiv:2010.11929}.

\bibitem[{Fu et~al.(2024)Fu, Chen, Shen, Qin, Zhang, Lin, Yang, Zheng, Li, Sun, Wu, and Ji}]{mme}
Fu, C.; Chen, P.; Shen, Y.; Qin, Y.; Zhang, M.; Lin, X.; Yang, J.; Zheng, X.; Li, K.; Sun, X.; Wu, Y.; and Ji, R. 2024.
\newblock MME: A Comprehensive Evaluation Benchmark for Multimodal Large Language Models.
\newblock arXiv:2306.13394.

\bibitem[{Goyal et~al.(2017)Goyal, Khot, Summers-Stay, Batra, and Parikh}]{vqa}
Goyal, Y.; Khot, T.; Summers-Stay, D.; Batra, D.; and Parikh, D. 2017.
\newblock Making the v in vqa matter: Elevating the role of image understanding in visual question answering.
\newblock In \emph{Proceedings of the IEEE conference on computer vision and pattern recognition}, 6904--6913.

\bibitem[{Grattafiori et~al.(2024)Grattafiori, Dubey, Jauhri, Pandey, Kadian, Al-Dahle, Letman, Mathur, Schelten, Vaughan et~al.}]{llama3}
Grattafiori, A.; Dubey, A.; Jauhri, A.; Pandey, A.; Kadian, A.; Al-Dahle, A.; Letman, A.; Mathur, A.; Schelten, A.; Vaughan, A.; et~al. 2024.
\newblock The llama 3 herd of models.
\newblock \emph{arXiv preprint arXiv:2407.21783}.

\bibitem[{Hudson and Manning(2019)}]{gqa}
Hudson, D.~A.; and Manning, C.~D. 2019.
\newblock Gqa: A new dataset for real-world visual reasoning and compositional question answering.
\newblock In \emph{Proceedings of the IEEE/CVF conference on computer vision and pattern recognition}, 6700--6709.

\bibitem[{Jiang et~al.(2024)Jiang, Sablayrolles, Roux, Mensch, Savary, Bamford, Chaplot, Casas, Hanna, Bressand et~al.}]{mixtral}
Jiang, A.~Q.; Sablayrolles, A.; Roux, A.; Mensch, A.; Savary, B.; Bamford, C.; Chaplot, D.~S.; Casas, D. d.~l.; Hanna, E.~B.; Bressand, F.; et~al. 2024.
\newblock Mixtral of experts.
\newblock \emph{arXiv preprint arXiv:2401.04088}.

\bibitem[{Kembhavi et~al.(2016)Kembhavi, Salvato, Kolve, Seo, Hajishirzi, and Farhadi}]{ai2d}
Kembhavi, A.; Salvato, M.; Kolve, E.; Seo, M.; Hajishirzi, H.; and Farhadi, A. 2016.
\newblock A diagram is worth a dozen images.
\newblock In \emph{Computer Vision--ECCV 2016: 14th European Conference, Amsterdam, The Netherlands, October 11--14, 2016, Proceedings, Part IV 14}, 235--251. Springer.

\bibitem[{Li et~al.(2023{\natexlab{a}})Li, Li, Savarese, and Hoi}]{blip2}
Li, J.; Li, D.; Savarese, S.; and Hoi, S. 2023{\natexlab{a}}.
\newblock Blip-2: Bootstrapping language-image pre-training with frozen image encoders and large language models.
\newblock In \emph{International conference on machine learning}, 19730--19742. PMLR.

\bibitem[{Li et~al.(2024)Li, Yuan, Liu, Tang, Wang, Zhu, and Zhang}]{tokenpacker}
Li, W.; Yuan, Y.; Liu, J.; Tang, D.; Wang, S.; Zhu, J.; and Zhang, L. 2024.
\newblock TokenPacker: Efficient Visual Projector for Multimodal LLM.
\newblock \emph{arXiv preprint arXiv:2407.02392}.

\bibitem[{Li et~al.(2023{\natexlab{b}})Li, Du, Zhou, Wang, Zhao, and Wen}]{pope}
Li, Y.; Du, Y.; Zhou, K.; Wang, J.; Zhao, W.~X.; and Wen, J.-R. 2023{\natexlab{b}}.
\newblock Evaluating object hallucination in large vision-language models.
\newblock \emph{arXiv preprint arXiv:2305.10355}.

\bibitem[{Lin et~al.(2023)Lin, Ye, Zhu, Cui, Ning, Jin, and Yuan}]{video-llava}
Lin, B.; Ye, Y.; Zhu, B.; Cui, J.; Ning, M.; Jin, P.; and Yuan, L. 2023.
\newblock Video-llava: Learning united visual representation by alignment before projection.
\newblock \emph{arXiv preprint arXiv:2311.10122}.

\bibitem[{Lin et~al.(2024)Lin, Lin, Lin, and Ji}]{VTW}
Lin, Z.; Lin, M.; Lin, L.; and Ji, R. 2024.
\newblock Boosting Multimodal Large Language Models with Visual Tokens Withdrawal for Rapid Inference.
\newblock \emph{arXiv preprint arXiv:2405.05803}.

\bibitem[{Liu et~al.(2024{\natexlab{a}})Liu, Li, Li, and Lee}]{llava1.5}
Liu, H.; Li, C.; Li, Y.; and Lee, Y.~J. 2024{\natexlab{a}}.
\newblock Improved baselines with visual instruction tuning.
\newblock In \emph{Proceedings of the IEEE/CVF Conference on Computer Vision and Pattern Recognition}, 26296--26306.

\bibitem[{Liu et~al.(2024{\natexlab{b}})Liu, Li, Li, Li, Zhang, Shen, and Lee}]{llava-next}
Liu, H.; Li, C.; Li, Y.; Li, B.; Zhang, Y.; Shen, S.; and Lee, Y.~J. 2024{\natexlab{b}}.
\newblock Llava-next: Improved reasoning, ocr, and world knowledge.

\bibitem[{Liu et~al.(2024{\natexlab{c}})Liu, You, Han, Liu, Huang, He, and Yang}]{visual_anchor}
Liu, H.; You, Q.; Han, X.; Liu, Y.; Huang, H.; He, R.; and Yang, H. 2024{\natexlab{c}}.
\newblock Visual Anchors Are Strong Information Aggregators For Multimodal Large Language Model.
\newblock \emph{arXiv preprint arXiv:2405.17815}.

\bibitem[{Liu et~al.(2025)Liu, Tang, Chen, Dong, Li, Zhou, Li, Hu, and Chu}]{can}
Liu, X.; Tang, Z.; Chen, H.; Dong, P.; Li, Z.; Zhou, X.; Li, B.; Hu, X.; and Chu, X. 2025.
\newblock Can LLMs Maintain Fundamental Abilities under KV Cache Compression?
\newblock \emph{arXiv preprint arXiv:2502.01941}.

\bibitem[{Liu et~al.(2023)Liu, Duan, Zhang, Li, Zhang, Zhao, Yuan, Wang, He, Liu et~al.}]{mmbench}
Liu, Y.; Duan, H.; Zhang, Y.; Li, B.; Zhang, S.; Zhao, W.; Yuan, Y.; Wang, J.; He, C.; Liu, Z.; et~al. 2023.
\newblock Mmbench: Is your multi-modal model an all-around player?
\newblock \emph{arXiv preprint arXiv:2307.06281}.

\bibitem[{Liu et~al.(2024{\natexlab{d}})Liu, Wu, Li, Tang, and Li}]{retrieval}
Liu, Y.; Wu, F.; Li, R.; Tang, Z.; and Li, K. 2024{\natexlab{d}}.
\newblock Retrieval Replace Reduction: An effective visual token reduction method via semantic match.
\newblock \emph{arXiv preprint arXiv:2410.07278}.

\bibitem[{Lu et~al.(2022)Lu, Mishra, Xia, Qiu, Chang, Zhu, Tafjord, Clark, and Kalyan}]{SQA}
Lu, P.; Mishra, S.; Xia, T.; Qiu, L.; Chang, K.-W.; Zhu, S.-C.; Tafjord, O.; Clark, P.; and Kalyan, A. 2022.
\newblock Learn to explain: Multimodal reasoning via thought chains for science question answering.
\newblock \emph{Advances in Neural Information Processing Systems}, 35: 2507--2521.

\bibitem[{Menendez et~al.(1997)Menendez, Pardo, Pardo, and Pardo}]{jensen}
Menendez, M.~L.; Pardo, J.; Pardo, L.; and Pardo, M. 1997.
\newblock The jensen-shannon divergence.
\newblock \emph{Journal of the Franklin Institute}, 334(2): 307--318.

\bibitem[{Oren et~al.(2024)Oren, Hassid, Adi, and Schwartz}]{tova}
Oren, M.; Hassid, M.; Adi, Y.; and Schwartz, R. 2024.
\newblock Transformers are multi-state rnns.
\newblock \emph{arXiv preprint arXiv:2401.06104}.

\bibitem[{Plummer et~al.(2015)Plummer, Wang, Cervantes, Caicedo, Hockenmaier, and Lazebnik}]{flickr30k}
Plummer, B.~A.; Wang, L.; Cervantes, C.~M.; Caicedo, J.~C.; Hockenmaier, J.; and Lazebnik, S. 2015.
\newblock Flickr30k entities: Collecting region-to-phrase correspondences for richer image-to-sentence models.
\newblock In \emph{Proceedings of the IEEE international conference on computer vision}, 2641--2649.

\bibitem[{Pope et~al.(2023)Pope, Douglas, Chowdhery, Devlin, Bradbury, Heek, Xiao, Agrawal, and Dean}]{kvcache}
Pope, R.; Douglas, S.; Chowdhery, A.; Devlin, J.; Bradbury, J.; Heek, J.; Xiao, K.; Agrawal, S.; and Dean, J. 2023.
\newblock Efficiently scaling transformer inference.
\newblock \emph{Proceedings of Machine Learning and Systems}, 5: 606--624.

\bibitem[{Radford et~al.(2021)Radford, Kim, Hallacy, Ramesh, Goh, Agarwal, Sastry, Askell, Mishkin, Clark et~al.}]{CLIP}
Radford, A.; Kim, J.~W.; Hallacy, C.; Ramesh, A.; Goh, G.; Agarwal, S.; Sastry, G.; Askell, A.; Mishkin, P.; Clark, J.; et~al. 2021.
\newblock Learning transferable visual models from natural language supervision.
\newblock In \emph{International conference on machine learning}, 8748--8763. PMLR.

\bibitem[{Shang et~al.(2024)Shang, Cai, Xu, Lee, and Yan}]{purmaerge}
Shang, Y.; Cai, M.; Xu, B.; Lee, Y.~J.; and Yan, Y. 2024.
\newblock Llava-prumerge: Adaptive token reduction for efficient large multimodal models.
\newblock \emph{arXiv preprint arXiv:2403.15388}.

\bibitem[{Shi et~al.(2023)Shi, Tao, Rao, Yang, Yuan, and Wang}]{crossget}
Shi, D.; Tao, C.; Rao, A.; Yang, Z.; Yuan, C.; and Wang, J. 2023.
\newblock Crossget: Cross-guided ensemble of tokens for accelerating vision-language transformers.
\newblock \emph{arXiv preprint arXiv:2305.17455}.

\bibitem[{Singh et~al.(2019)Singh, Natarajan, Shah, Jiang, Chen, Batra, Parikh, and Rohrbach}]{textvqa}
Singh, A.; Natarajan, V.; Shah, M.; Jiang, Y.; Chen, X.; Batra, D.; Parikh, D.; and Rohrbach, M. 2019.
\newblock Towards vqa models that can read.
\newblock In \emph{Proceedings of the IEEE/CVF conference on computer vision and pattern recognition}, 8317--8326.

\bibitem[{Tu et~al.(2024)Tu, Vashchilenko, Lu, and Xu}]{vl-cache}
Tu, D.; Vashchilenko, D.; Lu, Y.; and Xu, P. 2024.
\newblock VL-Cache: Sparsity and Modality-Aware KV Cache Compression for Vision-Language Model Inference Acceleration.
\newblock \emph{arXiv preprint arXiv:2410.23317}.

\bibitem[{Vaswani et~al.(2017)Vaswani, Shazeer, Parmar, Uszkoreit, Jones, Gomez, Kaiser, and Polosukhin}]{attention}
Vaswani, A.; Shazeer, N.; Parmar, N.; Uszkoreit, J.; Jones, L.; Gomez, A.~N.; Kaiser, L.; and Polosukhin, I. 2017.
\newblock Attention is all you need.
\newblock \emph{Advances in neural information processing systems}, 30.

\bibitem[{Xing et~al.(2024)Xing, Huang, Dong, Lu, Zhang, Zang, Cao, He, Wang, Wu et~al.}]{pyramiddrop}
Xing, L.; Huang, Q.; Dong, X.; Lu, J.; Zhang, P.; Zang, Y.; Cao, Y.; He, C.; Wang, J.; Wu, F.; et~al. 2024.
\newblock PyramidDrop: Accelerating Your Large Vision-Language Models via Pyramid Visual Redundancy Reduction.
\newblock \emph{arXiv preprint arXiv:2410.17247}.

\bibitem[{Yang et~al.(2024)Yang, Yang, Hui, Zheng, Yu, Zhou, Li, Li, Liu, Huang, Dong, Wei, Lin, Tang, Wang, Yang, Tu, Zhang, Ma, Xu, Zhou, Bai, He, Lin, Dang, Lu, Chen, Yang, Li, Xue, Ni, Zhang, Wang, Peng, Men, Gao, Lin, Wang, Bai, Tan, Zhu, Li, Liu, Ge, Deng, Zhou, Ren, Zhang, Wei, Ren, Fan, Yao, Zhang, Wan, Chu, Liu, Cui, Zhang, and Fan}]{qwen2}
Yang, A.; Yang, B.; Hui, B.; Zheng, B.; Yu, B.; Zhou, C.; Li, C.; Li, C.; Liu, D.; Huang, F.; Dong, G.; Wei, H.; Lin, H.; Tang, J.; Wang, J.; Yang, J.; Tu, J.; Zhang, J.; Ma, J.; Xu, J.; Zhou, J.; Bai, J.; He, J.; Lin, J.; Dang, K.; Lu, K.; Chen, K.; Yang, K.; Li, M.; Xue, M.; Ni, N.; Zhang, P.; Wang, P.; Peng, R.; Men, R.; Gao, R.; Lin, R.; Wang, S.; Bai, S.; Tan, S.; Zhu, T.; Li, T.; Liu, T.; Ge, W.; Deng, X.; Zhou, X.; Ren, X.; Zhang, X.; Wei, X.; Ren, X.; Fan, Y.; Yao, Y.; Zhang, Y.; Wan, Y.; Chu, Y.; Liu, Y.; Cui, Z.; Zhang, Z.; and Fan, Z. 2024.
\newblock Qwen2 Technical Report.
\newblock \emph{arXiv preprint arXiv:2407.10671}.

\bibitem[{Yao et~al.(2024)Yao, Li, Ren, Wang, Liu, Sun, and Hou}]{Deco}
Yao, L.; Li, L.; Ren, S.; Wang, L.; Liu, Y.; Sun, X.; and Hou, L. 2024.
\newblock DeCo: Decoupling Token Compression from Semantic Abstraction in Multimodal Large Language Models.
\newblock \emph{arXiv preprint arXiv:2405.20985}.

\bibitem[{Yue et~al.(2024)Yue, Ni, Zhang, Zheng, Liu, Zhang, Stevens, Jiang, Ren, Sun et~al.}]{mmmu}
Yue, X.; Ni, Y.; Zhang, K.; Zheng, T.; Liu, R.; Zhang, G.; Stevens, S.; Jiang, D.; Ren, W.; Sun, Y.; et~al. 2024.
\newblock Mmmu: A massive multi-discipline multimodal understanding and reasoning benchmark for expert agi.
\newblock In \emph{Proceedings of the IEEE/CVF Conference on Computer Vision and Pattern Recognition}, 9556--9567.

\bibitem[{Zhang, Li, and Bing(2023)}]{video-llama}
Zhang, H.; Li, X.; and Bing, L. 2023.
\newblock Video-llama: An instruction-tuned audio-visual language model for video understanding.
\newblock \emph{arXiv preprint arXiv:2306.02858}.

\bibitem[{Zhang et~al.(2024)Zhang, Fan, Ma, Zheng, Huang, Cheng, Gudovskiy, Okuno, Nakata, Keutzer et~al.}]{sparsevlm}
Zhang, Y.; Fan, C.-K.; Ma, J.; Zheng, W.; Huang, T.; Cheng, K.; Gudovskiy, D.; Okuno, T.; Nakata, Y.; Keutzer, K.; et~al. 2024.
\newblock SparseVLM: Visual Token Sparsification for Efficient Vision-Language Model Inference.
\newblock \emph{arXiv preprint arXiv:2410.04417}.

\bibitem[{Zhuang et~al.(2025)Zhuang, Hu, Mu, Hu, Liang, Ye, and Hu}]{falip}
Zhuang, J.; Hu, J.; Mu, L.; Hu, R.; Liang, X.; Ye, J.; and Hu, H. 2025.
\newblock FALIP: Visual Prompt as Foveal Attention Boosts CLIP Zero-Shot Performance.
\newblock In \emph{European Conference on Computer Vision}, 236--253. Springer.

\bibitem[{Zhuang et~al.(2024)Zhuang, Lu, Dai, Hu, Chen, Liu, and Hu}]{st3}
Zhuang, J.; Lu, L.; Dai, M.; Hu, R.; Chen, J.; Liu, Q.; and Hu, H. 2024.
\newblock ST3: Accelerating Multimodal Large Language Model by Spatial-Temporal Visual Token Trimming.
\newblock \emph{arXiv preprint arXiv:2412.20105}.

\end{thebibliography}
